\pdfoutput=1

\documentclass[11pt, dvipsnames]{article}

\usepackage[]{ACL2023}
\usepackage{graphicx}
\usepackage{svg}
\usepackage{changepage}
\usepackage{times}
\usepackage{latexsym}
\usepackage{amsfonts}

\usepackage[T1]{fontenc}
\usepackage{amsmath}
\usepackage[utf8]{inputenc}

\usepackage{microtype}

\usepackage{inconsolata}
\usepackage{breqn}
\usepackage{breqn}
\title{
 Could Chemical Language Models benefit from Message Passing}

\author{Jiaqing Xie, Ziheng Chi\\
  Department of Computer Science\\
  ETH Zurich \\
  \texttt{\{jiaxie, zihchi\}@student.ethz.ch} \\}

\begin{document}
\maketitle
\begin{abstract}

Pretrained language models (LMs) showcase significant capabilities in processing molecular text, while concurrently, message passing neural networks (MPNNs) demonstrate resilience and versatility in the domain of molecular science. Despite these advancements, we find there are limited studies investigating the relationship between molecular structures and their corresponding textual representations. Therefore, in this paper, we propose two strategies to evaluate whether an information integration can enhance the performance: contrast learning, which involves utilizing an MPNN to supervise the training of the LM, and fusion, which exploits information from both models. Our empirical analysis reveals that the integration approaches exhibit superior performance compared to baselines when applied to smaller molecular graphs, while these integration approaches do not yield performance enhancements on large scale graphs. Furthermore, we conduct experiments to assess the impact of dataset splitting strategies and random seeds on the overall performance. 

\end{abstract}

\section{Introduction}
The success of attention mechanisms on sequential data has introduced a massive family of large language models based on Transformer architecture \cite{vaswani2017attention}. It is evident that these large language models are useful for encoding sequential objects such as text \cite{liu2019roberta},  molecules \cite{honda2019smiles}, speech \cite{huang2021speech}, and forecasting data \cite{giuliari2021transformer}. It has been demonstrated that pretrained molecule language models are capable of encoding chemical elements semantically without learning structures \cite{honda2019smiles, xia2022mole, chithrananda2020chemberta, wang2019smiles}. Especially for proteins which function as natural components of the human body and a representative of molecule family, they could be efficiently encoded by transformer \cite{rao2019evaluating, elnaggar2021prottrans, rives2021biological, he2021pre} which acts as masked language modelers.

In contrast to text, molecules contain inherent relationships between their elements, indicating that structural encoding is necessary in addition to word embeddings. Message passing neural network (MPNN), emerging as a prominent method for encoding structural information in recent years, has demonstrated its robustness and versatility within the field of molecular sciences. By leveraging the 2-dimensional topological and 3-dimensional geometrical information as augmented features \cite{liu2021pre, stark20223d}, it is possible to learn molecular embeddings from structures without sequentially encoding traditional SMILES expressions. 

The advent of MPNNs has promoted the exploration of graph-based learning methods for molecular science. Graph contrastive learning captures potential different structural distributions to finetune self-learned representations, where both local and global features are enhanced with chemical domain expertise \cite{stark20223d, you2021graph, wang2022molecular}. Besides, the success of GPT \cite{radford2018improving} in traditional natural language processing tasks also motivates the research on graph transformers and graph GPT tailored for the molecule domain \cite{hu2020gpt, bagal2021molgpt, rong2020self, ying2021transformers, zhu2022featurizations}. 
\begin{figure}[!ht]
    \centering
    \includegraphics[width=0.43\textwidth]{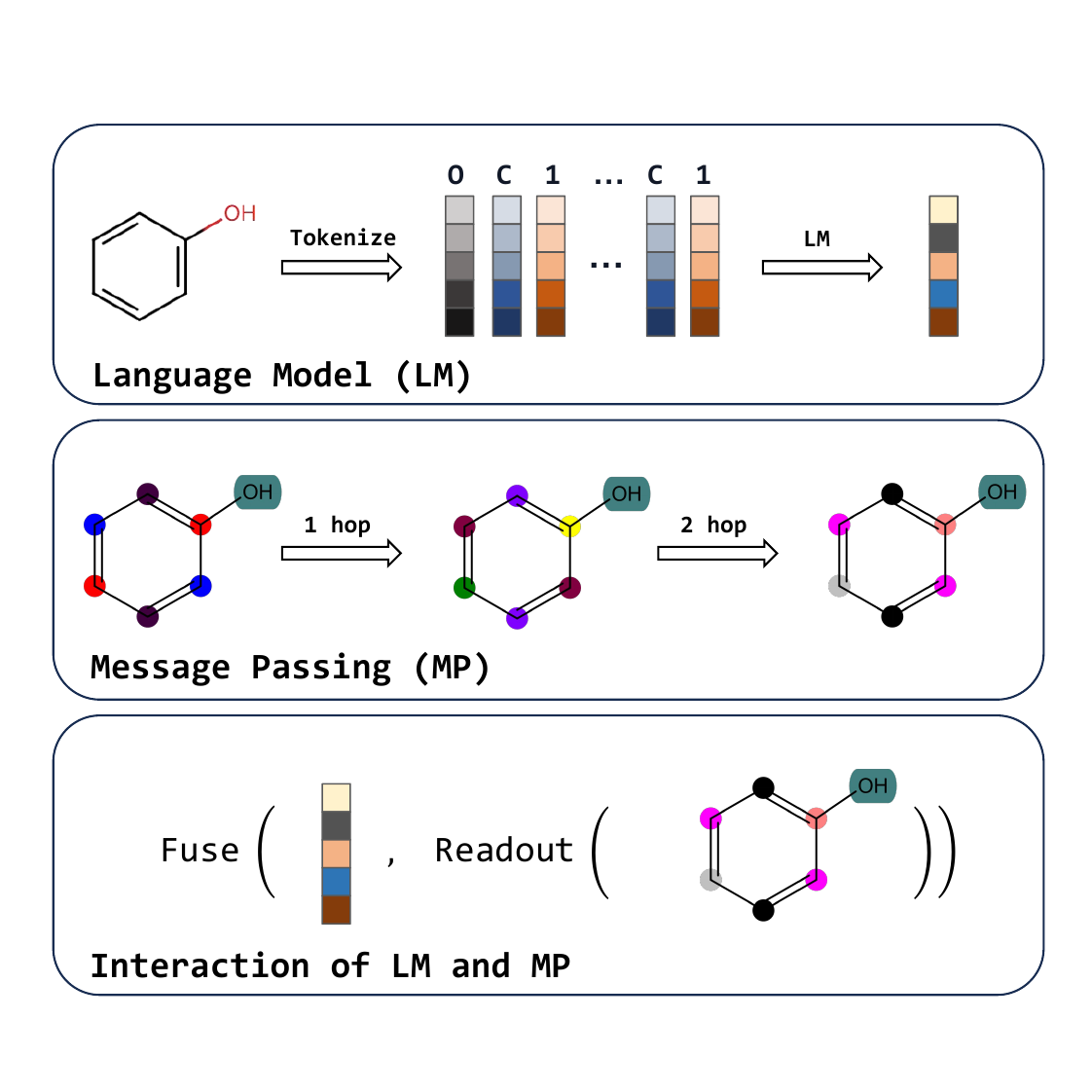}
    \vspace{-10pt}
    \caption{Baseline model: language model (LM) and message passing networks (MPNN). An interaction of LM and MPNN is investigated in this research.}
    \label{idea}
\end{figure}
Few studies has been investigated to appropriately merge text embeddings and graph embeddings for learning molecule representation better. There has been one study which demonstrated such relationship but with additional prompting with GPT model \cite{chen2024exploring}, which is out of our scope. In this paper, we aim to explore the interplay between molecular graph embeddings and SMILE token embeddings. We propose two categories of techniques for integrating information: contrast learning and fusion. In contrast learning-based methods, we incorporate an MPNN as an auxiliary model to supervise the training of the language model, operating at node or graph levels, while we only utilize the language model for downstream tasks. In fusion-based methods, we exploit information from both models to generate outputs for downstream tasks. This is achieved either by merging the output embeddings from both models or by integrating the output embeddings from one model with the input embeddings of the other. 

Our main contributions are to as follows: 
\begin{enumerate}
    \item Explore various information integration approaches to assess the necessity of incorporating supplementary structural features in molecular LLMs research, instead of pursuing state-of-the-art performance. 
    \item Benchmark a series of combination of sequential-based methods (LM) and structural-based methods (MPNN) as baselines for further research. 
\end{enumerate}

\section{Related Work}
\subsection{Molecule Representation Learning}
The Simplified Molecular Input Line Entry System (SMILES) has become a cornerstone in cheminformatics, providing a compact and standardized representation for chemical structures. Conserving molecular structural information and atom orderings, the SMILES descriptor converts a molecule from its structural representation into a condensed 1-dimensional textual sequence. For example, a phenol molecule ($\texttt{C}_6 \texttt{H}_5 \texttt{OH}$) is represented as $\texttt{C1=CC=C(C=C1)O}$. Similar to the tokenization in natural language settings, a molecule is expressed as a sentence and atoms are expressed as words. This allows efficient utilization of large language models in chemical research.

\subsection{Pretrained Large Language Models}
The advent of the transformer architecture \cite{vaswani2023attention} represents a breakthrough in the field of natural language processing. Over the past few years, many excellent pretraining strategies have been proposed, such as BERT \cite{devlin2019bert} and RoBERTa \cite{liu2019roberta}, significantly improving the capabilities of the large language models. As SMILES allows converting molecular structures into textual sentences, it is possible to apply language models for molecular machine learning, which facilitates the research on pretraining molecular language models. 

Based on the implementation of RoBERTa, ChemBERTa \cite{chithrananda2020chemberta} employs chemistry oriented masked-language modelling as its pretraining strategy, while the improved version ChemBERTa-2 \cite{ahmad2022chemberta2} adopts multi-task regression as another pretraining task and uses larger training datasets. There are also other BERT-like transformer models, such as MolBERT \cite{fabian2020molecular} and SMILES-BERT \cite{10.1145/3307339.3342186}, which are pretrained with different objectives on different molecule datasets.

\subsection{Contrastive Learning}
Contrastive learning has emerged as a powerful paradigm in self-supervised learning. Unlike traditional methods that rely solely on labeled data, this approach leverages the differences between data to learn representations. Based on the assumption that similar instances should be closer in the embedding space, the objective is to maximize the similarity between positive data pairs while minimizing the similarity between negative data pairs. So far, contrastive learning has demonstrated efficacy across diverse domains. A common practice of this approach is based on data augmentation \cite{you2021graph}, where the utilization of unlabeled data enhances model generalizability and robustness. Furthermore, this approach is also widely adopted in the field of multimodality \cite{radford2021learning}, where the availability of different data forms allows leveraging one representation to supervise the other. 

\begin{figure*}[!ht]
    \centering
    \includegraphics[width=0.7\textwidth]{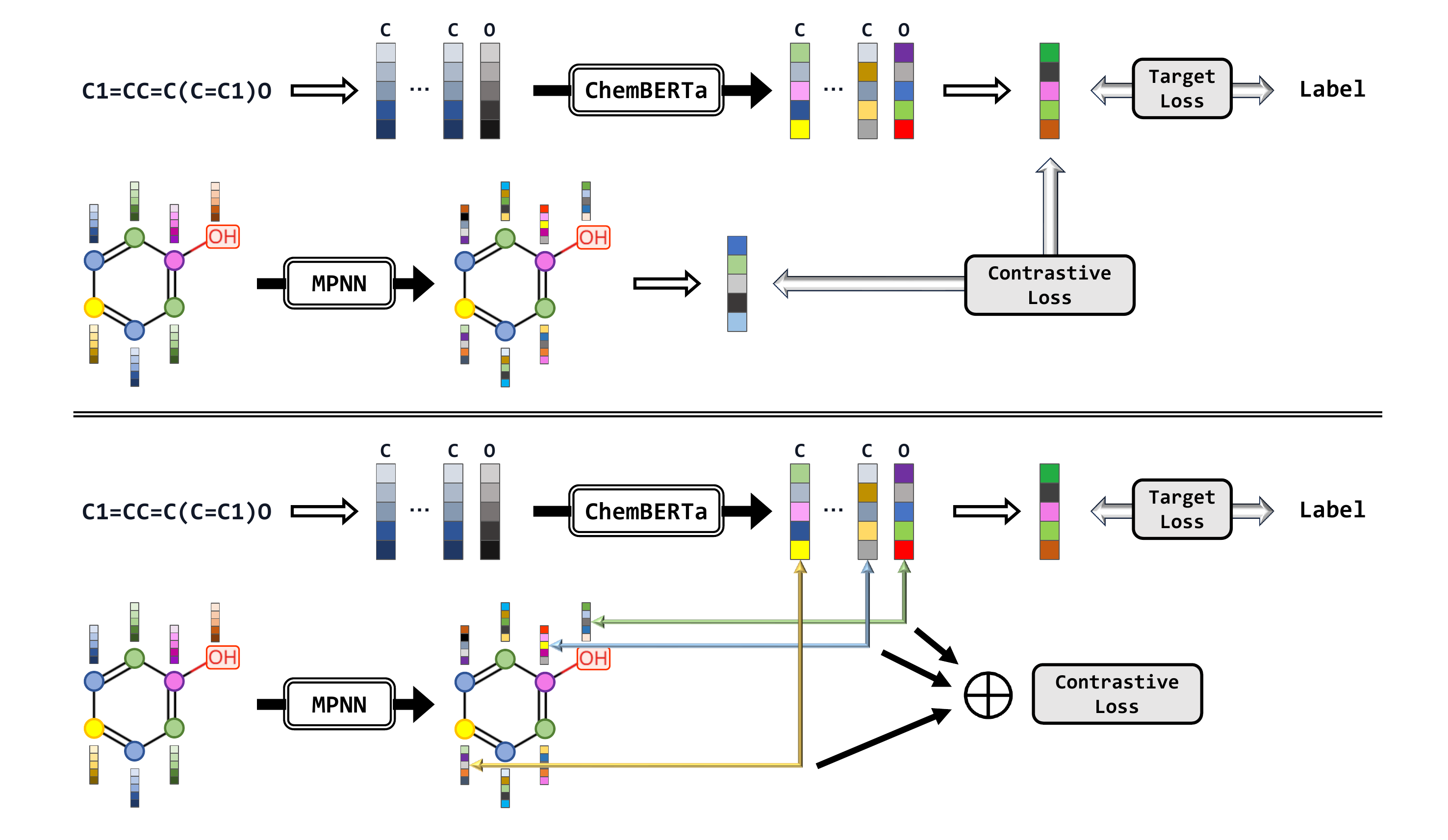}
    \caption{Contrastive Learning. Above: Node level contrastive learning. Below: Graph level contrastive learning. }
\end{figure*}

\section{Methods}
We investigate two kinds of information merging methods: contrast learning-based methods and fusion-based methods. In contrast learning-based methods, we use GNN as an auxiliary model to supervise the training of the language model, while we only use the language model for downstream tasks. In fusion-based methods, we make use of information from both models to generate the output for downstream tasks. For each kind of method, we consider different model architectures. In this section, we will first briefly review the two baseline models ChemBERTa \cite{chithrananda2020chemberta} and NNConv \cite{gilmer2017neural}, and then describe the contrast learning-based methods and the fusion-based methods.

\subsection{Baseline (Fig. 1)}

\paragraph*{Language Models. }
We choose ChemBERTa \cite{chithrananda2020chemberta} as our baseline model. The architecture of ChemBERTa is similar to BERT \cite{devlin2019bert}, consisting of an embedding layer and several encoder layers. A molecule is first converted into the textual format through SMILES, and then a SMILES tokenizer is applied to convert the words into input tokens. After embedding lookup, each token is assigned with an embedding. Then the encoder layers which consist of a multi-head self-attention layer and a feed-forward layer transform the input token embeddings to hidden state representations. Finally, the task-specific output layer (classifier or regressor) predicts the result. 

We ask the ChemBERTa model to produce three-level information for a molecule. First, node embeddings are extracted from the final hidden state representations. Since the SMILES transformation preserves atom orderings, each atom in the original molecule corresponds to a specific output token embedding. Second, the graph embedding is extracted from the special token at the beginning of the sequence. Third, the property prediction result is the final output. 

The entire process can be depicted as follows: 
\begin{align}
& \texttt{Tokens}=\texttt{Tokenizer}(\texttt{Sequence}) \notag \\
& E_{\text{in}}=\texttt{Embedding}(\texttt{Tokens}) \notag \\
& E_{\text{out}}=\texttt{Encoder}(E_{\text{in}}) \notag \\
& N=E_{\text{out}}[\texttt{Node\_Indices}] \notag \\
& G=E_{\text{out}}[0] \notag \\
& P=\texttt{Predictor}(G) \notag \\
\end{align}
\label{eq:ChemBERTa baseline}
where $E_{\text{in}}$ and $E_{\text{out}}$ represent the input token embeddings and final hidden state representations; $N$, $G$, and $P$ represent the node embeddings, graph embedding, and property prediction result. 


\paragraph*{Message Passing Neural Networks. }
There are different types of graph neural networks, which include graph convolution, graph attention and neural message passing networks (MPNN). Edge attributes or edge features are important in message passing mechanisms \cite{johannes2020directional, gilmer2017neural}. For the baseline model, we follow the model setting in the first paper of MPNN for Quantum Chemistry dataset QM9 \cite{gilmer2017neural}, which iteratively updates the message $ m^{t}_v$ and the hidden state $ h^{t}_v$ for each node v:
\begin{align}
    m^{(t+1)}_v &= \sum_{u \in \mathcal{N}(v)}\texttt{Aggr}[h^{(t)}_v, h^{(t)}_u, e_{uv}]\\
    h^{(t+1)}_v &= \mathcal{U}(h^{(t)}_v, m^{(t+1)}_v  )
\end{align} 
$\texttt{Aggr}[\cdot]$ is the function that aggregates neighbor node $u$'s information as well as the attributes $e_{uv}$ of the shared edge
with $u$. $\mathcal{U}( \cdot, \cdot)$ updates hidden states for $v$.

\subsection{Integration 1: Contrastive Learning (Fig. 2) }
\paragraph*{Node Level Contrastive Learning} In order to compute the constrastive loss, we need a triple $\langle \textbf{anchor, positive, negative} \rangle$. Anchor and positive node embeddings are sampled from language models and message passing networks respectively. Negative samples are randomly generated from graphs with a different permutation.
For an example node triple $t = \langle a, p, n \rangle$, its corresponding contrastive learning loss triplet loss is given by:
\begin{dmath}
L(t) = \max\{ d(a, p) - d(a, n) + \text{margin}, 0 \}
\end{dmath} where margin is 1.0 and distance measurement $d(i, j)$ is defined as $L_p$-norm : $d(i, j) = \| i - j\|_p$. p is often set to 2 as an Euclidean distance metric.
In a $\mathcal{M}$ mini-batch of training graphs (number of $\mathcal{N}$ nodes) with $\mathcal{K}$ triples, the triplet loss is given by:
\begin{dmath}
\small
\text{L} = \sum_{m = 1}^{\mathcal{M}} \sum_{i = 0}^{\lceil \frac{\mathcal{N}}{\mathcal{K}} \rceil - 1}\sum_{j = 1}^{\mathcal{K}} \max\{ d(a^m_k, p^m_k) - d(a^m_k, n^m_k) + \text{margin}, 0 \}
\end{dmath}
 where $k = |\mathcal{K}|i + j$. Consider a binary classification problem, as mentioned we need to perform $\texttt{Readout}$ function to obtain the global information of a graph. If it is an average function, the total loss is given by a prediction loss such as negative log likelihood (NLL) loss, and a regularized triplet contrastive loss which has been defined above: 
\begin{dmath}
    \small
    \text{L} =\sum_{m = 1}^{\mathcal{M}} \text{NLL}\left(\text{MLP}\left(\frac{1}{\mathcal{N}}\sum_{i = 0}^{\lceil \frac{\mathcal{N}}{\mathcal{K}} \rceil - 1}\sum_{j = 1}^{\mathcal{K}} a^m_k\right), y^m\right) + \alpha \cdot \sum_{m = 1}^{\mathcal{M}} \sum_{i = 0}^{\lceil \frac{\mathcal{N}}{\mathcal{K}} \rceil - 1}\sum_{j = 1}^{\mathcal{K}} \max\{ d(a^m_k, p^m_k) - d(a^m_k, n^m_k) + \text{margin}, 0 \}
\end{dmath} where $k = |\mathcal{K}|i + j$ likewise and $\alpha$ is a regularization term.

\begin{figure*}[!ht]
    \centering
    \includegraphics[width=0.7\textwidth]{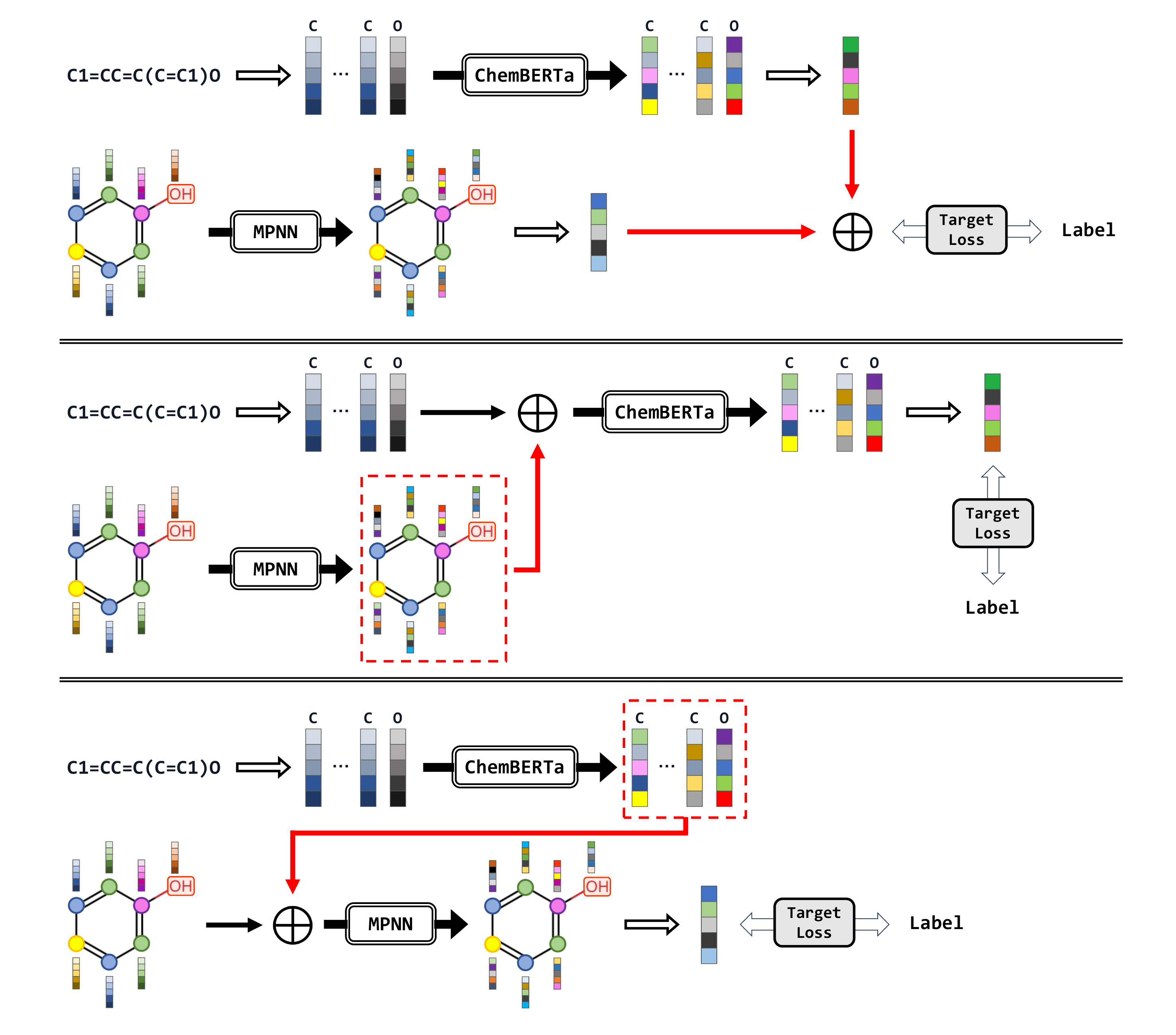}
    \caption{Fusion. Top: Late Fusion. Middle: \texttt{MPNN2LM} Joint Fusion. Bottom: \texttt{LM2MPNN} Joint Fusion. }
\end{figure*}

\paragraph*{Graph Level Contrastive Learning. }
Apart from establishing negative samples between each pair of nodes at a fine grained level, we estimate contrastive learning at a coarse grained level which aims at computing the difference between language model graph embeddings and MPNN graph embeddings. This could potentially avoid the situation that individual nodes with large difference contribute more to the difference of the molecule property. Moreover, the complexity is pretty lower than the complexity of node level comparison, which will be discussed in part 3.4. Similar to the node level training loss in (6), the graph level training loss is defined as:
\vspace{-1.2em}
\begin{dmath}
    \small
    \text{L} =\sum_{m = 1}^{\mathcal{M}} \text{NLL}\left(\text{MLP}\left( a^m\right), y^m\right) + \alpha' \cdot \sum_{m = 1}^{\mathcal{M}}  \max\{ d(a^m, p^m) - d(a^m, n^m) + \text{margin}, 0 \}
\end{dmath} where  $\alpha'$ is a regularization term.
Note that different molecules could have a similar graph embedding which could lead to a similar quantum property, for example isomers. And also note that we use message passing network outputs to self-supervise (or fine-tune) language model outputs either in node level and graph level settings. It means that we do not directly use MPNN outputs to perform predictions. This is because we want to see if injecting geometry information of molecules is beneficial to the end-to-end training of \textbf{language models}. It is different from the collaborative training (fusion) which would be introduced in the following parts.

\subsection{Integration 2: Fusion (Fig. 3)}
\paragraph*{Late Fusion} Different from self-supervised learning settings in part 3.2, we introduce another important interaction between LM embeddings and MPNN embeddings: late fusion \cite{sachan2021syntax}. It is called late fusion since the interaction happens after their corresponding embeddings $h_{\text{LM}}$ and $h_{\text{MPNN}}$ are extracted. The interaction is given by the notation $\bigoplus$, and the prediction is then given as:
\begin{align}
    y_{\text{pred}} &= \text{MLP}\left(h_{\text{LM}} \bigoplus h_{\text{MPNN}} \right) \\
    \bigoplus &= \{ +,  \max,  \Vert, \odot \}\\
    L &= \sum_{i = 1}^{\mathcal{M}}\text{NLL}(y^{i}_{\text{pred}}, y^{i})
\end{align}
where the interaction contains element-wise addition,  maximization, concatenation and gate function which is used for building highway layers. 
\paragraph*{Joint Fusion. }
Based on the late fusion, we consider two situations: 1) \texttt{MPNN2LM}: fuse initial graph embeddings and LM outputs $\Rightarrow$ perform MPNN downstream tasks, and 2)  \texttt{LM2MPNN}: fuse initial token embeddings and MPNN outputs $\Rightarrow$ finetune LM downsteam tasks.

\paragraph*{MPNN2LM}
We initialized the word embedding for language model: $h^{(0)}_{\text{LM}}$. MPNN embedding is given by $h_{\text{MPNN}}$. Then the fused embedding is $h^{(0)}_{\text{LM}} \bigoplus h_{\text{MPNN}}$, which would be the input embedding for the pretrained language model. The node mask is also considered since in part 3.1 we mentioned that paddings are added to ensure the same length of input. After fine-tuning pretrained \textbf{LM}, we readout the global information $h'$ to perform downstream tasks. Overall, the graph embedding prepared for MLP is:
\begin{equation}
    h' = \texttt{Readout}\left(\textbf{LM}\left(h^{(0)}_{\text{LM}} \bigoplus h_{\text{MPNN}}, \textbf{mask}\right)\right)
\end{equation}

\paragraph*{LM2MPNN}
Revisiting how MPNN works by (1) and (2), the node embedding for v are fused with the LM output for mask index v: $h^{(t)}_v \bigoplus h_{\text{LMM}(v)}$. Similar to the neighbor node u, they are fused with  the LM output for mask index u: $h^{(t)}_u  \allowbreak \bigoplus h_{\text{LMM}(u)}$. Then the message is aggregated by: 
\begin{dmath}
    m^{(t+1)}_v = \sum_{u \in \mathcal{N}(v)}\texttt{Aggr} \left(h^{(t)}_v \bigoplus h_{\text{LMM}(v)}, \allowbreak h^{(t)}_u  \allowbreak \bigoplus h_{\text{LMM}(u)}, \allowbreak e_{uv}\right)\\
\end{dmath}
\vspace{-1em}
For the update function $\mathcal{U}(\cdot, \cdot)$, the new update rule:
\begin{dmath}
        h^{(t+1)}_v = \mathcal{U} \left(h^{(t)}_v \bigoplus h_{\text{LMM}(v)}, \allowbreak m^{(t+1)}_v \right)
\end{dmath}

\subsection{Complexity Analysis}
\paragraph{Baseline Models. } The time complexity for baseline models are mainly dominated by their corresponding model architecture. Assume the input size $\in \mathbb{R}^{N x d}$
For the pretrained transformer model, the self-attention module is the bottleneck, which is bounded by $O(N^2 \cdot d)$. Assume that there are L self-attention layers, then it will increase to $O(N^2 \cdot d \cdot L)$. The complexity for feed forward layers is $O(N \cdot d^2 \cdot L)$. The overall complexity for baseline LM is then 
 $O(N^2 \cdot d \cdot L + N \cdot d^2 \cdot L)$. For MPNN, computing each message has a complexity of $O(d)$. The total complexity for the message passing step is then $O(E \cdot d)$. Updating nodes will be $O(N \cdot d^2)$. L layers lead to $O(L \cdot E \cdot d + L \cdot N \cdot d^2)$. So it depends on whether the graph is sparse or not. If graph is sparse, $N^2 \gg E$, then the complexity of LM is greater than the complexity of MPNN. This situation often occurs in real world applications.

\paragraph{Contrastive Learning. } For contrastive learning, apart from the basic complexity for LM and MPNN, it also includes the complexity for computing contrastive loss. Take triplet loss as an example, in equation (5), the complexity is dominated by the last term. Assume that computing max only requires O(1). Computing $d( \cdot, \cdot)$ requires $O(d)$ if the dimension of inputs is d. Then the overall complexity is $O(d \cdot  N)$ where N is the number of nodes in the graph. A node level contrastive learning then requires $O(N^2 \cdot d \cdot L + 2 \cdot N \cdot d^2 \cdot L + d\cdot N + L \cdot d \cdot E )$. For graph level contrastive learning we find that only complexity of model terms dominates, which leads to $O(N^2 \cdot d \cdot L + 2 \cdot N \cdot d^2 \cdot L +L \cdot d \cdot E + d)$, which is faster than that of node level.

\paragraph{Fusion}
We simply investigate $\bigoplus$ by choosing $\max$ which requires $O(1)$. The element wise maximization requires $O(N \cdot d)$ since input size is $N \times d$. Then the time complexity of late fusion would be the same as the complexity of nodel level contrastive learning, which is $O(N^2 \cdot d \cdot L + 2 \cdot N \cdot d^2 \cdot L +L \cdot d \cdot (N + E) )$. \texttt{MPNN2LM} has the same complexity while \texttt{LM2MPNN} is much more complex since $\bigoplus$ directly affects the complexity of message passing operation. We already know that the complexity of MPNN is $O(E \cdot d)$. We assume that the average number of nodes is $\frac{2E}{N}$. Then the additional element wise addition contributions additional $O(E \cdot d)$, which leads to the overall complexity for \texttt{LM2MPNN}: $O(N^2 \cdot d \cdot L + 2 \cdot N \cdot d^2 \cdot L +L \cdot d \cdot (N + 2E) )$.

\begin{table*}[!htb] \small
\centering
\begin{tabular}{lcccc}
\hline
Model & HIV $(\textbf{acc.})\uparrow$  & BACE $(\textbf{acc.})\uparrow$ & BBBP $(\textbf{acc.})\uparrow$  & ESOL $(\textbf{mae.})\downarrow$ \\  \hline
ChemBERTa      &  0.9776 $\pm$ 0.0021  &  $0.8280 \pm 0.0319$     &   $0.9105 \pm 0.0153$    &   $0.5529 \pm 0.0332$   \\
MPNN  &  0.9774 $\pm$ 0.0022   &  0.8080 $\pm$ 0.0256  &  0.8737 $\pm$ 0.0140   &  0.6252 $\pm$ 0.0072     \\\hline
ChemBERTa  contra.  MPNN (node)      &  \textbf{0.9782} $\pm $ \textbf{0.0035}   & $0.8280 \pm 0.0271$     & 0.9118 $\pm$  0.0245 &   $0.5326  \pm 0.0534$  \\
ChemBERTa  contra.  MPNN (graph)      &   0.9774 $\pm$ 0.0022  &   $0.8300 \pm 0.0352$    &  $0.9131 \pm 0.0307$   &  $0.5404 \pm 0.0495$   \\
ChemBERTa  + MPNN (graph) &  0.9778 $\pm$ 0.0020&  $0.8320  \pm 0.0331$     &  0.9065 $\pm$ 0.0147 & 0.5002 $\pm$ 0.0339 \\
ChemBERTa $\leftarrow$ MPNN  &   0.9773 $\pm$ 0.0022  & 0.8060 $\pm$ 0.0422      &  $0.9053 \pm 0.0099$     & $\textbf{0.4819} \pm \textbf{0.0325}$    \\
ChemBERTa $\rightarrow$ MPNN   &  0.9773 $\pm$ 0.0022  &  $\textbf{0.8380}  \pm \textbf{0.0366}$    & $\textbf{0.9184} \pm \textbf{0.0189} $     &  0.5561 $\pm$ 0.0461 \\\hline 
\end{tabular}
\caption{Performance of pretrained ChemBERTa on MoleculeNet datasets. }
\end{table*}

\begin{table*}[htb] \small
\centering
\begin{tabular}{lcccc}
\hline
Model & HIV $(\textbf{acc.})\uparrow$  & BACE $(\textbf{acc.})\uparrow$ & BBBP $(\textbf{acc.})\uparrow$  & ESOL $(\textbf{mae.})\downarrow$ \\  \hline
ChemBERTa-2   &  $0.9792 \pm 0.0018$   &   $0.8560 \pm 0.0206$  &   $0.9171 \pm 0.0136$    &     0.4738 $\pm$ 0.0330 \\
MPNN   &   0.9774 $\pm$ 0.0022   &  0.8010 $\pm$ 0.0392  &  0.8737 $\pm$ 0.0140  &   0.6252 $\pm$ 0.0072    \\\hline
ChemBERTa-2 contra. MPNN (node)      & $0.9791 \pm 0.0011$   &  0.8620 $\pm$ 0.0256 &  $\textbf{0.9290} \pm \textbf{0.0128}$   &  \textbf{0.4393} $\pm$ \textbf{0.0338} \\
ChemBERTa-2  contra. MPNN (graph)      &  $\textbf{0.9800} \pm \textbf{0.0017}$  & 0.8540 $\pm$ 0.0258    & $0.9197 \pm 0.0163$     &  0.4643 $\pm$ 0.0354  \\
ChemBERTa-2  + MPNN (graph)      &   $0.9791 \pm 0.0012$ & $\textbf{0.8680} \pm \textbf{0.0293}$  &  $0.9263 \pm  0.0113$   & $0.4493 \pm 0.0328$  \\
ChemBERTa-2 $\leftarrow$ MPNN &  $ 0.9772 \pm 0.0016$  &    0.8400 $\pm$ 0.0374 &  $0.8974 \pm 0.0098$  & 0.5012 $\pm$ 0.0335    \\
ChemBERTa-2 $\rightarrow$ MPNN  &  $0.9789 \pm 0.0012$  &   $0.8480 \pm 0.0204$   &   $0.9224 \pm 0.0141$  &  $0.4516 \pm  0.0264$ \\\hline
\end{tabular}
\caption{Performance of improved pretrained ChemBERTa-2 on MoleculeNet datasets. }
\end{table*}

\section{Experiment settings}
\subsection{Dataset}

We follow the following paradigm \cite{luo2022one} for prediction on quantum chemistry based datasets: first we perform tests on small scale and classifical benchmark molecule datasets. In our future works,  we want to test its robustness on large scale and recently proposed benchmarks such as  PCQM4Mv2 \cite{hu2020open}. For small datasets we choose from MoleculeNet dataset \cite{wu2018moleculenet} which  collects data from physical chemistry, biophysics and physiology field. It has provided plenty of molecule datasets to play with \cite{wu2018moleculenet}. For large datasets, we choose \textbf{QM9} \cite{gilmer2017neural} as tested in MPNN. The task is to predict property for each molecule using models in part 3. Selected datasets are \textbf{HIV}, \textbf{BACE},  \textbf{ESOL} and \textbf{BBBP} \cite{wu2018moleculenet}. A simple description of chosen dataset and task type is listed in table 3.
\begin{table}[!htp] \small
\centering
\begin{tabular}{lccccc}
\hline
\textbf{Name} & \textbf{\#graphs} & \textbf{\#nodes} & \textbf{\#features} & \textbf{\#classes} \\
\hline
HIV & 41,127 & $\sim$25.5 & 9 & 1 \\
BBBP & 2,050 & $\sim$23.9 & 9 & 1 \\
BACE & 1,513 & $\sim$34.1 & 9 & 1 \\
ESOL & 1,128 & $\sim$13.3 & 9 & 1 \\
QM9 & 130,831 & $\sim$18.0 & 11 & 19 \\
\hline
\end{tabular}
\caption{Descriptions of selected datasets from MoleculeNet}
\end{table}
\textbf{HIV}, \textbf{BBBP}, and \textbf{BACE} are used for binary classification settings, while \textbf{ESOL} and \textbf{QM9} are used for regression settings. For simplicity, we only choose the first target from all 19 classes, which is the Dipole moment $\mu$. For the regression problem, the performance is measured by mean absolute error (mae). As for the classification, it is measured by the mean accuracy (acc).  Specifically, the pretrained ChemBERTa is time-consuming on QM9 and HIV dataset.

\subsection{Hyper-parameter settings}
There are two pretrained model to choose from: ChemBERTa and its improved version ChemBERTa-2. We choose Adam optimizer for optimizing model parameters with default
learning rate 0.001 when running with pretrained ChemBERTa-2 (<4G). The initial learning rate is tuned to 0.0002 when running with   ChemBERTa since the model size is large (>16G) which requires a small learning rate. We follow a 8:1:1 train-valid-test ratio for MoleculeNet dataset, and follow an approximate 21:2:2 train-valid-test ratio for \textbf{QM9} dataset. Hidden dimension is set to 64. The default choice for $\bigoplus$ is sum (addition). Five fixed seeds are 0, 7, 42, 100, 2024 for result reproduction.

\subsection{Scalability} A single NVIDIA A100 GPU could satisfy all our experiments. In other words, it is scalable for training all datasets including large scaled ones. The maximum usage is observed when running pretrained ChemBERTa on HIV dataset. For other datasets it's also possible to train on a GeForce RTX 3090 GPU.

\section{Results}

\paragraph*{Observation 0: Protein language models are more preferred.} A fundamental observation from experimenting on MoleculeNet is that purely using message passing neural networks are inferior to language models in molecule property prediction. This phenomenon is also mentioned in the  previous research work \cite{xu2022peer}. This has indicated some works to include the geometric properties such as 3D information and rotation invariant parameters in message passing networks to reinforce its prediction and expressive power. The explanation of this phenomenon would be that 1)  model size of either ChemBERTa-1 or ChemBERTA-2 model is larger than the size of message passing networks and 2)  either ChemBERTa-1 or ChemBERTA-2 model has been pretrained on some more larger datasets for example ZINC dataset, while message passing networks do not follow the pretraining scheme of large language models. 

\paragraph*{Observation 1: Integration on relatively small graphs are more preferred. } 
Using the pretraind ChemBERTa-2, we found that both contrastive learning and fusion methods outperform baseline models in \textbf{ESOL}, \textbf{BACE}, and \textbf{BBBP} where they are relatively small compared with \textbf{QM9} and \textbf{HIV} datasets. Especially, node level contrastive learning performs the best and it seems to be robust among all tasks, followed by late fusion methods and joint fusion methods when injecting LLM to MPNNs. In large dataset, the tuning strategy might influence the potential performance, where it splits the dataset in a better way therefore we perform one ablation regarding train test split (in section 6)
to avoid the difference that brought by dataset itself. 

\paragraph*{Observation 2: Integration w.r.t both regression and classification are useful. } 
In terms of training convergence, we observe that the accuracy or mean absolute error converges quickly to a high or low score respectively. For small graph datasets \textbf{BACE} and \textbf{BBBP} on graph classification problem, an improvement of \textbf{$\approx$1\%} on average accuracy is observed with method \texttt{MPNN2LM} for pretrained ChemBERTa. For version 2, $1.4 \%$ improvement is observed with late fusion on \textbf{BACE} and  $1.3 \%$ improvement is observed  with node contrastive learning on \textbf{BBBP} . For small graph dataset \textbf{ESOL} on regression problem, a great improvement is observed where $12.8 \%$ improvement on mae with MPNN2LM method with pretrained ChemBERTa, and $7.3 \%$ improvement on mae with MPNN2LM method with pretrained ChemBERTa-2. For \textbf{HIV}, we observe a little improvement with node level contrastive learning. Using a combination of LLM representation and graph representation during the training would make the prediction worse. For \textbf{QM9}, most of the injection / fusion methods would potentially improve the performance except for MPNN2LM fusion. Using LM2MPNN would potentially improve 8.6 \%. We found that pure MPNN's performance is better than the performance of a chemical LLM (table 4).

\paragraph*{Observation 3: Pretrained language models are important for downstream predictions. }
In comparison to ChemBERTa-2, ChemBERTa performs worse when comparing each entry in table 1 and table 2.  Although we could always try to improve those two baselines with different injection or fusion methods, the best of them are not the same. For example, contrastive learning is much more preferred to ChemBERTa-2 while fusion methods are much more preferred to ChemBERTa model. When it comes with a new pretrained large language model, using our proposed method could tell the similarity between tasks and the model's pre-training strategy. As there is no general conclusion about how a chemical LLM and a MPNN could be combined to predict the best, it is still a pioneering area that requires more pretrained models to test its robustness.
To select the most appropriate pretrained language model for further training, researchers should first integrate a list of pretrained models, followed by an investigation with different fusion / injection methods.

\paragraph*{Observation 4: Joint Fusion to some extent helps learn MPNN better but learn original Chemical LLM worse. }
We also focus on if such multi-modal module (Fig. 2, Fig. 3) helps learn individual module (Fig. 1) better. It improves a lot for single MPNN baseline if we consider its language level information as augmented features. For example, for \textbf{BACE} dataset, MPNN has an average accuracy of 0.808 with ChemBERTa. With injecting pretrained language information, an improvement of 3.7\% is observed (LM2MPNN). However, it might not work very well on the opposite when we inject information from MPNN to LLM. For simplicity we just examined with pretrained ChemBERTa-2. For ESOL dataset, it decreased from 0.4738 to  0.5012 (5.78\%). For BBBP dataet, it decreased from 0.9171 to 0.8974 (2.15\%).
We further suggest that the researchers should not directly use the structural information from graphs as additional input when they want to modify their LLM models, but trying to leverage them as auxiliary ground-truth to finetune the token embeddings.

\begin{table}[!htb] \small
\centering
\begin{tabular}{lc}
\hline
Model & QM9 (target = 0) \\  \hline 
ChemBERTa-2 baseline &  0.4825 $\pm$  0.0113 \\
MPNN baseline    &  0.4669 $\pm$ 0.0065\\\hline
ChemBERTa-2 contra. MPNN (node) & 0.4613 $\pm$ 0.0065 \\
ChemBERTa-2 contra. MPNN (graph) & 0.4662 $\pm$ 0.0046 \\
ChemBERTa-2 + MPNN (graph)   &  0.4596 $\pm$ 0.0078 \\
ChemBERTa-2 $\leftarrow$ MPNN & 0.5231 $\pm$ 0.0083    \\
ChemBERTa-2 $\rightarrow$ MPNN  &   \textbf{0.4409} $\pm$ \textbf{0.0048}  \\\hline
\end{tabular}
\caption{Performance of improved pretrained ChemBERTa-2 on QM9 dataset. }
\end{table}

\section{Ablation Study}
\paragraph{Effects of datasets. }
We choose another dataset in MoleculeNet to certify that the proposed models are still robust on this dataset. Take \textbf{FreeSolv} as an example, we figure out that none of the injection or contrastive learning methods is still robust on this regression task.
Even if late fusion performs the best which has an average mae of 0.6568, which is close to the result of pure chemical LLM training (0.6420), there's still a 2.3\% decrease in performance. Both \texttt{LM2MPNN} and \texttt{MPNN2LM} did not work well, but it still commits to our fourth main observation, which is that injecting token embeddings into message passing layers would still improve the performance, but injecting structural information into word embeddings would be a bad idea. A potential reason is that \textbf{FreeSolv} is too small.  We suggest that researchers should be careful  when fine-tuning the individual language model with additional structural features.

\begin{table}[!htb] \small
\centering
\begin{tabular}{lc}
\hline
Model & FreeSolv $(\textbf{mae.})\downarrow$ \\  \hline 
ChemBERTa-2 baseline &  0.6420 $\pm$ 0.0814  \\
MPNN baseline    &   0.9904 $\pm$ 0.1375 \\ \hline
ChemBERTa-2 contra. MPNN (node) & 0.6642 $\pm$ 0.0600  \\
ChemBERTa-2 contra. MPNN (graph) & $ 0.6745  \pm 0.0995$ \\
ChemBERTa-2 + MPNN (graph)   &  $ 0.6568 \pm 0.0658$\\
ChemBERTa-2 $\leftarrow$ MPNN &   0.9188 $\pm$ 0.0686  \\
ChemBERTa-2 $\rightarrow$ MPNN  &  $0.7475 \pm 0.0805$  \\\hline
\end{tabular}
\caption{Performance of improved pretrained ChemBERTa-2 on FreeSolv dataset}
\end{table}
\paragraph{Effects of dataset split. }
We want to figure out if different splits of training, validation and test datasets lead to different performance. We run on \textbf{BBBP} (classification) and \textbf{ESOL} (regression). Four ratios are considered: 9:0.5:0.5, 8:1:1, 7:2:1, and 6:2:2. Model prediction power is highest at a ratio of 8:1:1 for \textbf{ESOL} while the prediction power is reducing for \textbf{BBBP} when ratio of training sets is decreasing.

\begin{table}[!htb] \small
\centering
\begin{tabular}{lcc}
\hline
Train test split & BBBP $(\textbf{acc.})\uparrow$ & ESOL $(\textbf{mae.})\downarrow$ \\  \hline 
9:  0.5 : 0.5 & \textbf{0.9500} $\pm$ \textbf{0.0174}& 0.4672 $\pm$ 0.0338 \\
8 : 1 : 1&  $0.9290 \pm 0.0128$ &  \textbf{0.4393} $\pm$ \textbf{0.0338}\\
7 : 2 : 1& 0.9211 $\pm$ 0.0110 & 0.4837 $\pm$ 0.0447\\
6 : 2 : 2& 0.9152 $\pm$ 0.0032&  0.4947 $\pm$ 0.0060\\
 \hline
\end{tabular}
\caption{Model (node level contrast.) }
\end{table}

\paragraph{Effects of different fusion operations $\bigoplus$} 
We first follow the default train valid test split of 8:1:1. As mentioned, there are four fusion operations $\bigoplus$, which are max, sum, concatenation and gate function. Our default fusion operation is sum function. Surprisingly we found that concatenation and max function are better fusion choice for both \textbf{BBBP} and \textbf{ESOL}. We suggest that  researchers could simply concatenate token embeddings and graph embeddings together. 

\begin{table}[!htb] \small
\centering
\begin{tabular}{lcc}
\hline
Fusion Operation & BBBP $(\textbf{acc.})\uparrow$ & ESOL $(\textbf{mae.})\downarrow$ \\  \hline 
sum & $0.9263 \pm  0.0113$  &$0.4493 \pm 0.0328$\\
max & 0.9289 $\pm$  0.0146 & 0.4281 $\pm$ 0.0339\\
concate & \textbf{0.9289} $\pm$  \textbf{0.0241} & \textbf{0.4255} $\pm$ \textbf{0.0335} \\
gate & 0.9224 $\pm$ 0.0197  & 0.4363 $\pm$ 0.0354\\ \hline
\end{tabular}
\caption{Model: Late Fusion  }
\end{table}

\paragraph{Effects of different graph neural networks}
As mentioned in section 3, there are three types of graph neural networks in mainstream GNN research, which are graph convolution (GraphConv), message passing neural networks (MPNN), and graph attention networks. We substitute MPNN with with a two-layer GraphConv model to see if MPNN is much better than other types of GNN for baselines. The results show that MPNN is more preferred to \textbf{BBBP} but GraphConv is more preferred to \textbf{ESOL}. Overall the difference would not be too large for a graph convolution network and a neural message passing layer therefore we suggest researchers try out both ways to improve the results.

\begin{table}[!htb] \small
\centering
\begin{tabular}{ccc}
\hline
Fusion Operation & BBBP $(\textbf{acc.})\uparrow$ & ESOL $(\textbf{mae.})\downarrow$ \\  \hline 
MPNN & \textbf{0.9289} $\pm$  \textbf{0.0146} & 0.4281 $\pm$ 0.0339\\
GraphConv & 0.9237 $\pm$  0.0148 & \textbf{0.4144} $\pm$ \textbf{0.0252}\\
\hline
\end{tabular}
\caption{Model: Late Fusion }
\end{table}

\section{Conclusion}
In this paper, we delved into various information integration approaches to assess whether the collaborative utilization of chemical large language models (chemical LLMs) and message passing neural networks (MPNNs) surpasses the individual efficacy of these models. We evaluated the integration approaches over different graph scales on both classification and regression tasks. Our empirical analysis has demonstrated that the integration approaches outperform the baselines on small-scale graphs but do not yield improvements on datasets of larger scales. Furthermore, we have found that differences in dataset splitting strategies,  and aggregation choices in fusion have an impact on the overall performance. We wish to extend our proposed methods on large scale benchmark datasets such as  PCQM4Mv2 \cite{hu2020open}.

\bibliography{custom}
\bibliographystyle{acl_natbib}
\appendix



\end{document}